\documentclass[a4paper,11pt,twocolumn,twoside]{article}
\usepackage[dvips]{graphicx}
\usepackage{sepln_en}
\usepackage{fullname}
\usepackage[hyphens]{url}
\usepackage[breaklinks]{hyperref}
\usepackage{graphicx}
\usepackage{xcolor}
\usepackage{multicol}
\usepackage{adjustbox}
\usepackage[utf8]{inputenc}
\usepackage[english]{babel}
\usepackage{multirow}
\usepackage{color}
\usepackage{latexsym}
\usepackage[T1]{fontenc}
\usepackage{microtype}
\usepackage{booktabs,colortbl,tabularx}
\usepackage{float}
\restylefloat{table}
\usepackage{url}
\usepackage{adjustbox}
\usepackage{makecell}

\input epsf

\title{Open Generative Large Language Models for Galician}

\author {\textbf{Pablo Gamallo$^1$,} \textbf{Pablo Rodríguez$^1$,} \textbf{Iria de-Dios-Flores$^2$,} \textbf{Susana Sotelo$^1$}, \\
\textbf{Silvia Paniagua$^1$,} \textbf{Daniel Bardanca$^1$,} \textbf{José Ramom Pichel$^1$,} \textbf{Marcos Garcia$^1$}  \\
$^1$ Centro Singular de Investigación en Tecnoloxías Intelixentes (CiTIUS) \\ Universidade de Santiago de Compostela\\
$^2$ Department of Translation and Language Sciences \\ Universitat Pompeu Fabra \\
\small{\texttt{\{pablo.gamallo,pablorodriguez.fernandez,susana.sotelo.docio\}@usc.gal}} \\ 
\small{\texttt{\{silvia.paniagua.suarez,danielbardanca.outeirino,jramon.pichel\}@usc.gal}}\\
\small{\texttt{marcos.garcia.gonzalez@usc.gal}, \texttt{iria.dedios@upf.edu}}
}

\seplntranstitle{Grandes Modelos de Lengua Generativos y Abiertos para Gallego} 

\seplnresumen{Los grandes modelos de lengua (LLM por su nombre en inglés) han transformado el procesamiento del lenguaje natural, pero la predominancia del uso de datos en inglés para su entrenamiento ha dado lugar a sesgos y disparidades de rendimiento entre lenguas. Este desequilibrio margina a las lenguas minoritarias, dificultando el acceso equitativo a las tecnologías de PLN para las lenguas con menos recursos, como el gallego. Para hacer frente a esta situación, presentamos los dos primeros LLM generativos centrados en el gallego. Estos modelos, disponibles gratuitamente como recursos de código abierto, han sido entrenados utilizando una arquitectura GPT con 1,3 mil millones de parámetros, a partir de un corpus de 2,1 mil millones de palabras.  Aprovechando la técnica de pre-entrenamiento continuado, hemos adaptado al gallego dos LLM existentes entrenados en corpus más grandes, mitigando así las limitaciones de datos que surgirían si el entrenamiento se realizara desde cero. Los modelos se han evaluado utilizando juicios humanos y conjuntos de datos basados en tareas de referencia estandarizadas. Estas evaluaciones revelan un rendimiento prometedor, subrayando la importancia de la diversidad lingüística en los modelos generativos.}
\seplnclave{grandes modelos de lengua, gallego, pre-entrenamiento continuado, acceso abierto.}

\seplnabstract{Large language models (LLMs) have transformed natural language processing. Yet, their predominantly English-centric training has led to biases and performance disparities across languages. This imbalance marginalizes minoritized languages, making equitable access to NLP technologies more difficult for languages with lower resources, such as Galician. We present the first two generative LLMs focused on Galician to bridge this gap. These models, freely available as open-source resources, were trained using a GPT architecture with 1.3B parameters on a corpus of 2.1B words.  Leveraging continual pretraining, we adapt to Galician two existing LLMs trained on larger corpora,  thus mitigating the data constraints that would arise if the training were performed from scratch. The models were evaluated using human judgments and task-based datasets from standardized benchmarks. These evaluations reveal a promising performance, underscoring the importance of linguistic diversity in generative models.}
\seplnkey{large language models, Galician, continual pertaining, open access.}

\firstpageno{1}

\begin{document}

\setlength\titlebox{21.5cm} 

\label{firstpage} \maketitle

%

\section{Introduction}


In recent years, large language models (LLMs) have revolutionized natural language processing by exhibiting remarkable capabilities in understanding and generating human-like text across various languages. These models, predominantly trained on vast English corpora, have become pivotal tools in multiple downstream tasks, ranging from machine translation to text summarization and sentiment analysis \cite{WangGLUE2018,PapernoLAMBADA2016}. Despite the large number of languages spoken worldwide, the predominant use of English text during training has resulted in these models exhibiting biases and disparities in performance across different languages. The vast majority of the training data is in a few dominant languages, namely English, with only a fraction dedicated to other languages, leaving under-resourced languages or varieties marginalized and underrepresented.\footnote{For instance, 92.65\% of the training data of GPT3 was English text (source: {\url{https://github.com/openai/gpt-3/blob/master/dataset_statistics/languages_by_word_count.csv}}).}
However, as the global landscape continues to evolve towards linguistic diversity, it is imperative to address the limitations inherent in current LLMs, particularly regarding their treatment of content in under-resourced languages. Furthermore, the lack of adequate representation for minority or minoritized languages hinders equitable access to NLP technologies and services for diverse linguistic communities, perpetuating the digital language divide and reinforcing linguistic hegemony \cite{Khanuja2023}.

To address this imbalance, this article presents the creation of the two first generative LLMs focused on the Galician language, a Romance language (also considered a variety of Portuguese) spoken primarily in the autonomous community of Galicia. By developing these specialized large decoder models, we hope that companies or third parties can benefit from them and integrate them into their applications, adapting them to the specific technological needs, and the linguistic needs and cultural context of Galician speakers. In alignment with the principles of open science, the models, associated datasets, and training corpora reported in this work are freely available as open-source resources. These models were created under the auspices of the \href{http://www.nos.gal}{Nós Project}, an initiative by the Universidade de Santiago de Compostela aimed at providing Galician with openly licensed resources and tools in the area of language-centric AI \cite{de-dios-flores-etal-2022-nos}. It is integrated with the \href{http://www.https://proyectoilenia.es/}{Ilenia Project}, which aims to generate digital resources that allow the development of multilingual applications in the different languages of Spain.

To achieve our objective, we explore a strategy based on continual pretraining, an efficient technique to build new LLMs \cite{Gupta2023}. The main motivation to follow this strategy stems from the fact that low-resource languages such as Galician face a major data constraint when adopting the common approach of pre-training from scratch with randomly initialized weights. The volume of data necessary for this undertaking can be enormous, rendering it unattainable. However, by commencing from a fully-trained LLM, we can leverage the existing knowledge encapsulated within it. To achieve this, only the weights of the embedding layer need adjustment. When provided with source and target vocabularies, we retain the weights corresponding to the shared tokens, while initializing the remainder as the average of the embeddings in the source vocabulary.

The strategy we follow in the current work to train the Galician LLMs consists of adapting by means of continual pretraining other existing models that contain a high proportion of Ibero-Romance languages to a greater or lesser extent close to Galician. Specifically, our Galician models are the result of adapting trilingual LLMs of Catalan, Spanish and English, which are, in turn, the result of another adaptation of foundational models with a majority presence of English. Our contribution focuses exclusively on the latter adaptation to Galician, being the intermediate model with the three mentioned languages a softened approximation of the starting foundational model that was trained with little or no text in the target language. The intermediate languages are, on the one hand, close languages to Galician (Spanish and Catalan), and on the other, the language that is best represented in the starting foundational model (English). In sum, this strategy was used to create two Galician LLMs, resulting from the continual pretraining of two existing trilingual models (Catalan/Spanish/English) built within the AINA project\footnote{\href{https://projecteaina.cat/}{https://projecteaina.cat/}} with continual pretraining from two foundation models based mainly on English, although one of them is multilingual.

The two Galician LLMs we have developed are evaluated in two different ways: 1) a systematic qualitative human evaluation is performed by identifying types of errors made when generating text from different contexts (or prompts); 2) a quantitative automatic evaluation is performed on several tasks by using common benchmarking datasets which we translated into Galician. In this second task-based evaluation, the performance of our Galician LLMs models is compared with multilingual LLMs.

 This paper is organized as follows. We begin in Section \ref{sec:related-work} by introducing the main LLMs within the framework of the Iberian languages. We continue in Section \ref{sec:method} by describing the methodology of the work, focusing on the continual pretraining strategy as well as on the characteristics of the corpus used in the training. Section \ref{sec:experiments} presents how the experiments for the training of the two models were carried out, while Section \ref{sec:evaluation} describes in detail the two types of evaluations performed. Finally, we discuss conclusions and future work in the last section. 

\section{Related Work: LLMs for Iberian languages} \label{sec:related-work}

Over the past few years, multilingual large language models have emerged as the prevailing method for constructing NLP and AI systems. However, it is worth emphasizing that these multilingual models have predominantly been developed for English. For instance, in Llama \cite{Touvron2023llama}, 89.7\% of the training data was in English,  with roughly 9\% in unknown languages. The remainder, which is a minuscule percentage of the training data, is reserved for a wide range of languages, including German (0.17\%), French (0.16\%), and Chinese (0.13\%). The presence of minority languages such as Galician is residual.

However, in the last years, there has been an effort to develop language models for Iberian and Romance languages which are worth noting. The AINA project has emerged as a significant contributor to this endeavor, offering pre-trained models such as FLOR-1.3B and FLOR-6.3B \cite{Flor2024}.\footnote{\href{https://medium.com/@mpamies247/flor-\-6-3b-a-chinchilla-compliant-model-for-catalan-spanish-and-english-7cdb389a9aac}{FLOR-6.3B: a chinchilla-compliant model for Catalan, Spanish and English}}
 These models, developed by the AINA research group, exhibit substantial capabilities in understanding and generating text in Catalan. It should be noted that one of our models (Carballo-bloom-1.3B below) is a continual pretraining from Flor-1.3B which, in addition to Catalan, was also trained with Spanish and English, in balanced percentages. More precisely, the training corpus of Flor-1.3B comprises 26B words in these three languages, with a smaller portion of English data.

In the domain of Spanish language processing, the MarIA family of both auto-encoding and generative models should be highlighted. This family, as presented in the study by \cite{GutierrezMarIA2022}, encompasses various models with different architectures (RoBERTa and GPT2), which were evaluated in different NLP tasks showcasing robust performance across a range of benchmarks. The models were pretrained using a massive corpus of 135B words extracted from the Spanish Web Archive crawled by the National Library of Spain between 2009 and 2019. 

For Portuguese, generative models have been explored extensively to capture the intricacies of the language. Notable works include those by \cite{LopesGlorIA2024} and \cite{SantosGervasio2024}, which describe the development of two generative models for Portuguese, Glória and Gervásio, respectively. Glória adopts the GPTNeo architectures with 1.3B and 2.7B parameters \cite{Hendrycks2021}. It must be emphasized that the Glória model was trained exclusively with Portuguese texts from Portugal (35B tokens), excluding Brazilian. Gervásio is a family of two 7B parameters models based on continual pretraining from LLaMA 2 \cite{Touvron2023llama}, one trained with Brazilian data and other with corpora from Portugal.

In relation to Galician, until now there was no large generative (autoregressive) language model. Two auto-encoding models are worth mentioning, namely Bertinho \cite{VilaresBertinho2021} and Bert-Galician \cite{Garcia2021}, both models with two versions: 6 (small) and 12 (base) transformer layers. 

Additionally, efforts to develop LLMs for the Basque language have yielded significant advancements. The Latxa models stand out as a notable example of autoregressive LLMs \cite{Latxa2024}.\footnote{\href{https://huggingface.co/HiTZ/latxa-7b-v1}{https://huggingface.co/HiTZ/latxa-7b-v1}} Two models were built by continual pretraining from LLaMA 2 with only 288M words ranging from 7B to 70B parameters, which are currently the biggest and best-performing LLMs built for Basque. Latxa models demonstrate good performance in various Basque language instruction tasks, contributing to the growing repertoire of NLP resources for Basque.


As can be observed, there is no clear methodology to build a LLM adapted to a particular language, since each project works with different architectures, different base models, and a training corpus size that depends on the available material for each language. Nonetheless, a common strategy in resource-poor contexts is the use of the continual pretraining strategy. In the following, we present our methodology and the characteristics of the Galician training corpus.

\section{Methodology} \label{sec:method}

\subsection{Continual pretraining}
Continual pretraining enables incremental learning on the same task, instead of retraining a model from scratch each time new data or a new language is introduced.  This can lead to more efficient use of computational resources and reduced training time, as the model only needs to learn from the new data. It is important to note that the data needed for pretraining foundational models increases proportionally with the number of parameters. Consequently, as model sizes expand, training a LLM from the scratch can become prohibitively expensive. This challenge is especially pronounced for languages with limited resources, where gathering the necessary data to effectively train billion-parameter models can be extremely challenging \cite{Gupta2023}.

The specific objective of the present work is to build a new open-source model that can understand and generate Galician language in a better way, by making use of continual pretraining on a multilingual base model. In addition to being more efficient in computational terms, continual pretraining allows the multilingual base model to adapt and learn the specific linguistic patterns and characteristics of the new language. This can be done because continual pretraining on a new language enhances the model's cross-lingual transfer capabilities, enabling it to transfer knowledge learned from the languages of the base model to the target language.  Knowledge transfer in continual pretraining can be applied not only across languages but also across domains to improve a model's performance in different tasks \cite{Ke2023}. As a result, knowledge transfer enables the target model to better understand and generate text in Galician language. 

The general method to perform continual pretraining is to use a base model that was trained using data from languages that are similar to Galician so that the final model can benefit from a non-random initialization of its weights, hence, requiring fewer new tokens. The initial stage of a successful language adaptation involves substituting the model's tokenizer. This step is pivotal because employing the original model tokenizer would result in a low proportion of token splits. Therefore, a new Byte Pair Encoding (BPE) tokenizer was trained using Galician text. Subsequently, the embedding layer undergoes modification by retaining solely the weights corresponding to shared tokens (those found in both the old and new tokenizer), while replacing the remaining ones with the overall mean value \cite{Downey2023}. Once the model is suitably initialized, standard pretraining procedures can start using our monolingual corpus.

\subsection{The corpus}
The two models described in this work have been trained with CorpusNÓS \cite{de2024corpusnos}, the largest collection of openly available
Galician texts. It is made up of 13.95GB of text (2.1B words) primarily devised for training large language models (LLMs). The corpus sources are varied and represent a relatively wide range of genres (see Table \ref{tab:corpus} for details). Crucially, the corpus is divided into two subcorpus depending on how the texts were obtained (either via transfer agreement from the text owners or from publicly available sources). A cleaning pipeline was developed, mainly based on perplexity to remove boilerplate and reduce noise in text \cite{Pyplexity2024}. CorpusNÓS, as well as the cleaning pipeline developed to process the texts, is made available via the project's official GitHub repository: \texttt{\url{https://github.com/proxectonos/corpora}}.

\begin{table*}[h!]
\centering
\caption{Corpus statistics.}
\label{tab:corpus}
\begin{tabular}{|c|l|r|r|}
\hline
\rowcolor[HTML]{D9D9D9} \textbf{Subcorpus} & \textbf{Genre} & \textbf{Nº tokens} & \textbf{Nº documents} \\ \hline
\multirow{7}{*}{\textbf{\makecell{1. Data obtained \\ via transfer agreement}}} & Books & 7.255.784 & 104 \\ \cline{2-4} 
 & Research articles & 2.665.351 & 664 \\ \cline{2-4} 
 & Press & 124.253.084 & 224.419 \\ \cline{2-4} 
 & Governmental & 245.897.880 & 654.505 \\ \cline{2-4} 
 & Web contents & 15.946.686 & 44.165 \\ \cline{2-4} 
 & Encyclopedic & 4.799.214 & 47.396 \\\cline{2-4}  
 & \multicolumn{1}{|r|}{\textbf{Subtotal}} & \textbf{400.817.999} & \textbf{971.253} \\ \specialrule{1.5pt}{1pt}{1pt}
\multirow{5}{*}{\textbf{2. Public data}} & Press and blog & 153.497.883 & 665.265 \\ \cline{2-4} 
 & Encyclopedic & 57.164.848 & 184.628 \\ \cline{2-4} 
 & Web crawls & 1.384.015.664 & 3.366.449 \\ \cline{2-4} 
 & Translation corpora & 133.726.004 & 4.745.799 \\ \cline{2-4}
 & \multicolumn{1}{|r|}{\textbf{Subtotal}} & \textbf{1.728.404.399} & \textbf{8.777.514} \\ \specialrule{1.5pt}{1pt}{1pt}
 \multicolumn{2}{|r|}{\textbf{Total}} & \textbf{2.129.222.398} & \textbf{9.748.767}  \\ \hline
\end{tabular}
\end{table*}

\section{Experiments} \label{sec:experiments}

In our experiments, we explore the efficacy of two distinct base models: FLOR-1.3B,\footnote{\href{https://huggingface.co/projecte-aina/FLOR-1.3B}{https://huggingface.co/projecte-aina/FLOR-1.3B}} built upon a multilingual Bloom 1.7B architecture,\footnote{\href{https://huggingface.co/bigscience/bloom-1b7}{https://huggingface.co/bigscience/bloom-1b7}} and Cerebras-GPT-1.3B, which follows the GPT-3 architecture and was initially pretrained solely in English\footnote{\href{https://huggingface.co/cerebras/Cerebras-GPT-1.3B}{https://huggingface.co/cerebras/Cerebras-GPT-1.3B}}. Both decoder architectures feature 16 attention heads and 24 layers, where the hidden layers have 2048 dimensions.
The choice of a monolingual and a multilingual model as start point aims to assess whether or not incorporating multiple languages in the initial training enhances performance during subsequent continual pretraining with Galician. It is important to note we do not use these models directly; instead, we initialize our training in versions already tailored to Spanish, English, and Catalan, as previously mentioned. As a result, we obtained two Galician LLMs: Carballo-bloom-1.3B,\footnote{\href{https://huggingface.co/proxectonos/Carballo-bloom-1.3B}{https://huggingface.co/proxectonos/Carballo-bloom-1.3B}} derived from Bloom 1.7B and a Catalan/Spanish/English continual pretraining, and Carballo-cerebras-1.3B,\footnote{\href{https://huggingface.co/proxectonos/Carballo-cerebras-1.3B}{https://huggingface.co/proxectonos/Carballo-cerebras-1.3B}}, derived from Cerebras-GPT-1.3B and an analogous trilingual continual pretraining.

To adapt the tokenizer model to the Galician language, a new Byte Pair Encoding (BPE)  tokenizer was trained on our corpus giving rise to a Galician vocabulary with 50257 tokens. To achieve the final adaptation, in the embedding layer, only the weights associated with shared tokens are retained, that is, we keep the weights of those tokens found in both the source Catalan/Spanish/English tokenizer and the Galician one. By contrast, the weights associated with the remainder tokens, namely those that are only in the Galician tokenizer, were substituted by the overall mean value computed over all tokens in the Catalan/Spanish/English tokenizer.

 To pretrain the models to Galician, we maintained consistent hyperparameters across both experiments, aligning closely with those employed in at least one of the models. For optimization, we employ Adam \cite{kingma2017} with $\beta_1=0.9$, $\beta_2=0.999$, and $\epsilon=10^{-8}$, coupled with a weight decay of $0.1$. The learning rate commences at $5\times10^{-5}$ and decays linearly. The sequence length is fixed at 2048 tokens, mirroring the Bloom and Cerebras base models. Training was performed with BF-16 mixed-precision.

We leverage the HuggingFace Transformers library \cite{wolf-etal-2020} for executing the Causal Language Modeling pre-training objective, while DeepSpeed \cite{rajbhandari2020} with ZeRO stage 2 optimizations is used to accelerate training. All the experiments were conducted utilizing nodes equipped with NVIDIA A100 40GB GPUs.
 All the experiments were conducted at the Galician Supercomputing Center (CESGA) utilizing nodes equipped with NVIDIA A100 40GB GPUs.

\section{Evaluation} \label{sec:evaluation}
Evaluating the performance of generative LLMs poses many challenges as their superficial linguistic abilities approximate that of humans and the kind of tasks they can perform are becoming increasingly diverse. One of the main challenges comes from the lack of a clear ground truth, as a given ``correct'' or ``appropriate'' answer may be phrased in a virtually infinite number of ways. This complicates the objective assessment of the quality of the generated text. Generative LLMs are commonly assessed through automated evaluation techniques, leveraging established datasets (i.e. benchmarks) that allow standardization and comparison across models. Nonetheless, the availability of these benchmarks in languages other than English is quite limited. Additionally, human evaluation provides valuable insights into quality and accuracy, drawing from expert and user assessments --although this approach incurs significant costs \cite{10.1145/3641289}. A hybrid approach strikes a balance between computational precision and human judgment. As we explain in what follows, our models have been evaluated using a mixture of human judgments and automated evaluation datasets from commonly used benchmarks that have been adapted to Galician.

\subsection{Qualitative human evaluation}
To measure the perceived quality of the models' outputs, particularly from a linguistic perspective, we devised a human perception experiment with expert linguists.  The evaluators were presented with a context and a continuation, and their task consisted of identifying the presence of errors of different categories (explained below) on the continuation. The continuation could either be synthetically generated text or authentic text, which was used to establish a baseline. The details of the methodology are explained below.

\subsubsection{Methodology}

\paragraph{Participants:}  6 expert linguists naive to the details of the task were selected for the evaluation of the two models. 

\paragraph{Materials:} The materials consisted of 60 texts extracted from CorpusNÓS before its use for training. These texts,  which were not part of the training,  were equally distributed along the different text types of the training corpus (e.g. novel, press, dialog, etc.), and were selected from paragraphs roughly having a maximum length of 1400 characters and a minimum of 250 (the resulting texts had a mean length of 644 characters). The texts were split in order to obtain a context and a continuation. To increase variability, we devised four splitting strategies: they could be split at the beginning or the middle of the paragraph (e.g. after the first sentence or after several sentences), and in the middle of a sentence or at the end of a sentence. These four configurations were counterbalanced. From the 60 texts selected, we created two versions for the continuations: the original text (i.e. authentic) and the synthetic text (i.e. the model's generation). The resulting 120 evaluation items were distributed in two lists (two sub-experiments) using a Latin Square design to ensure that no evaluator saw the same context in the two versions, and such that the two sub-experiments had the same number of authentic and synthetic texts. The length of the models' continuations for the synthetically generated texts was restricted to the length of the original text to avoid significant length differences between authentic and synthetic texts. An example of an evaluation context and the two possible continuations are provided below (an English translation is available in Appendix \ref{sec:anex}.)\footnote{Note that the generated texts may not end with a full sentence due to the length restrictions. The evaluators were instructed not to take this into account.}

\begin{itemize}
  \item \textbf{Context:} \textit{As universidades poderán contratar docentes de linguas modernas para os seus lectorados como parte do}
  \item \textbf{Authentic continuation:} \textit{persoal dos departamentos responsables. Os estatutos de cada universidade fixarán as características deste persoal. As súas obrigas docentes virán fixadas nos respectivos contratos, en virtude das necesidades docentes do departamento, sendo como máximo, as dun profesor colaborador.}
  \item \textbf{Generated continuation:} \textit{seu programa docente. A contratación destes profesores deberá ser aprobada pola universidade correspondente, que poderá establecer as condicións e requisitos específicos dos candidatos a profesor lector ou bolseiro en función das necesidades da institución académica na cal se vaia integrar o}
\end{itemize}

\paragraph{Procedure:} The evaluators were presented with a context and a continuation and were instructed to exclusively evaluate the continuation by indicating whether there were errors in the categories described below. The evaluation was binary, such that the presence of one or multiple errors would be considered erroneous.

\begin{itemize}
  \item \textbf{Form error:} The text is perfectly understandable but contains at least one formal error that can be identified by a non-expert reader.
  \item \textbf{Content error:} The text is well written but contains some incoherence of meaning, either with respect to the context or to the text itself, making it difficult to understand.
    \item \textbf{Register error:} The text does not adapt to the register (formal, informal, etc.) and/or to the typology of the genre (poetry, dialogue, narration, quotation, etc.) of the context, or the style changes abruptly. 
  \item \textbf{Repetitive content:} The text contains unjustifiable repetitions of words, segments, or content in relation to the context or the continuation. 
  \item \textbf{Inappropriate content:} The text contains pornographic, racist, hateful, sexist, or insulting language. 
  \item \textbf{Factual error:} The text provides incorrect information that can be objectively verified. Attention was paid only to information that can be considered to be known by a person with secondary education. When in doubt, evaluators were instructed to check only if the information could be quickly and easily consulted online.
\end{itemize}

The evaluators were provided with guidelines and examples before performing the evaluation and were asked to take as much time as needed. On average, the task lasted approximately 1h.

\subsubsection{Results and discussion}
\begin{figure*}[h!]
    \centering
    \includegraphics[width=\textwidth]{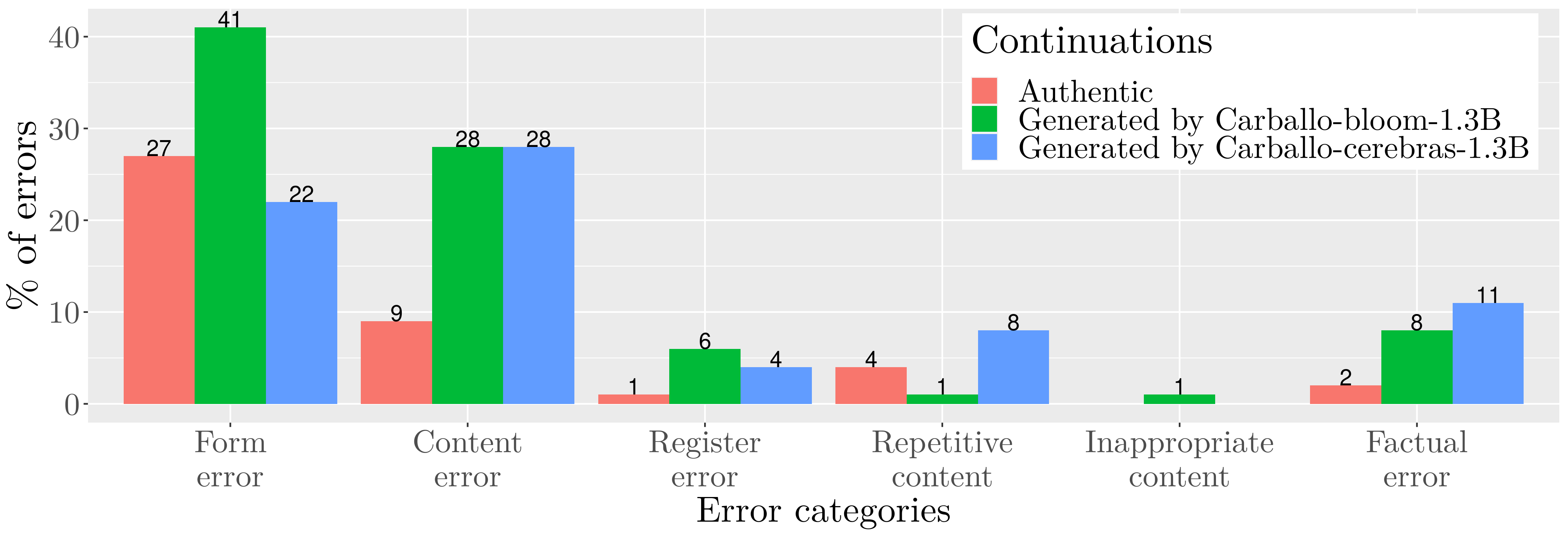}
    \caption{Human evaluation results.}
    \label{fig:1}
\end{figure*}

Figure \ref{fig:1} displays the results for the evaluations of the authentic texts and the continuations of those same texts produced by the two models. The graph shows, for each type of error, the percentage of continuations or authentic texts with at least one error of that type. It can be observed that the vast majority of errors identified were concentrated in the categories of form and content error, while the percentage of errors observed in the rest of the categories were residual. One striking result concerns the relatively high presence of form errors in authentic texts. Upon a detailed examination, it was identified that the vast majority of these errors were punctuation errors (e.g. ``,,'' instead of ``,'', ``.....'' instead of ``\ldots'', opening exclamation and question marks, extra spaces, wrong capitalizations, etc.). Importantly, none of the texts classified as having form errors contained grammatical or spelling errors. These same form errors observed in authentic texts are transferred and amplified by the models in their generations (particularly in Carballo-bloom-1.3B, where the errors are found in 41\% of continuations). Interestingly, however, Carballo-cerebras-1.3B generates better-formed text than the authentic one: 22\% vs. 27\%. We interpret these findings as an indication that the training corpus needs additional cleaning processes. Furthermore, while content errors are only marginally present in authentic texts (9\%), 28\% of errors were identified in the texts generated by the two models. Upon examination, it was determined that the vast majority of these errors had to do with abrupt topic changes. Nonetheless, it should be noted that the evaluators were quite strict in their judgments. Overall, it was observed that the vast majority of generated texts were grammatically correct and semantically coherent, while the vast majority of the errors identified by the evaluators were rather minor. 
The results also show that the two models perform similarly, although it is worth noting that, surprisingly, Carballo-cerebras-1.3B, built on an English-only model, performs slightly better than Carballo-bloom-1.3B, which is based on a multilingual foundation model, specifically in relation to form errors. 

While this evaluation provides valuable insights into the quality and coherence of texts generated by our models, it is limited by the selection of only 60 samples for each model and the involvement of 6 evaluators. Consequently, the findings should be interpreted with caution, and further research with larger sample sizes and more evaluators is warranted to establish more robust conclusions. Despite these limitations, our study serves as a stepping stone toward understanding the linguistic capabilities of our models, shedding light on both their strengths and areas for improvement.

\subsection{Task-based evaluation}

\begin{table*}[h!]
\centering
\caption{Task-based evaluation results. \textit{Bl-}, \textit{FL-}, and \textit{Cer-} refer to Bloom, FLOR, and Cerebras-GPT models, respectively.}
\begin{tabular}{| l | r r r r r |}
\hline
\multirow{2}{*}{\textbf{Model}} & \multicolumn{5}{c|}{\textbf{Task}}\\
\cline{2-6}
& \multicolumn{1}{c}{\it Belebele} & \multicolumn{1}{c}{\it CoLA} & \multicolumn{1}{c}{\it OpenBookQA} & \multicolumn{1}{c}{\it Parafrases-gl} & \multicolumn{1}{c|}{\it PAWS-X} \\
\hline
\hline
\textit{Carballo-bl} & 0.231±0.014 & 0.499±0.012 & 0.364±0.022 & 0.523±0.031 & \textbf{0.541±0.011} \\
\textit{Carballo-ce} & \textbf{0.271±0.015} & 0.502±0.012 & \textbf{0.368±0.022} & 0.496±0.031 & 0.531±0.011 \\
\hline
\textit{Bl-1b1} & 0.234±0.014 & \textbf{0.507±0.012} & 0.338±0.021 & 0.485±0.031 & 0.508±0.011 \\ 
\textit{Bl-1b7} & 0.218±0.014 & 0.500±0.012 & 0.338±0.021 & \textbf{0.539±0.031} & 0.539±0.011 \\ 
\textit{mGPT} & 0.229±0.014 & 0.494±0.012 & 0.332±0.021 & 0.423±0.031 & 0.517±0.011 \\
\textit{FL-1.3B} & 0.220±0.014 & 0.504±0.012 & 0.342±0.021 & 0.516±0.031 & 0.536±0.011 \\ 
\textit{Cer-1.3B} & 0.221±0.014 & 0.497±0.012 & 0.300±0.021 & 0.492±0.031 & 0.531±0.011 \\ 
\hline
\end{tabular}
\label{tab:taskeval}
\end{table*}

The task-based evaluation of the two models was carried out using EleutherAI's LM Evaluation Harness \cite{eval-harness} on the following five tasks, all of them with a 5-shot setup:


\begin{itemize}
    \item \textbf{Belebele}: Machine reading comprehension dataset available in over 100 languages, featuring questions based on short passages, each accompanied by four multiple-choice answers (only one correct) \cite{bandarkar2023belebele}.

    \item \textbf{OpenBookQA}: Question answering dataset combining multiple-choice questions on elementary-level science facts \cite{OpenBookQA2018}. It provides an assessment of both understanding and reasoning since it requires the combination of these scientific facts with common knowledge to answer the questions.

    \item \textbf{CoLA}: A two-choice dataset containing sentences labeled to assess linguistic acceptability \cite{warstadt-etal-2019-neural}.

    \item \textbf{Parafrases-gl}: Dataset with nearly 3000 entries labeled in three categories (paraphrases, non-paraphrases, and boundary paraphrases). It includes paraphrases at the syntactic level (generated by back-translation) and word level (produced by automatic substitution of terms).

    \item \textbf{PAWS-X}: Two-choice dataset for paraphrase detection in 6 languages, with high lexical overlap \cite{yang-etal-2019-paws}.
\end{itemize}

These datasets were converted by human translation into Galician, with the exception of \textit{parafrases-gl}, which was directly generated from Galician texts. They will be publicly released in the upcoming months, as part of a \textit{Galician Benchmark}.

For the task-based evaluation, we chose three multilingual open models of comparable sizes as baselines for comparison, as there are no other generative models for Galician. Bloom-1b1\footnote{\href{https://huggingface.co/bigscience/bloom-1b1}{https://huggingface.co/bigscience/bloom-1b1}} (1.1B)
and Bloom-1b7 
(1.7B) were trained with 350B unique tokens in 45 languages, while mGPT\footnote{\href{https://huggingface.co/ai-forever/mGPT}{https://huggingface.co/ai-forever/mGPT}} is a 1.3B multilingual model trained on 440B tokens in 61 languages. It's worth noting that these three models include Portuguese texts in their training corpora, but not Galician. Furthermore, the two models from which our models were derived via continual pretraining were also evaluated: FLOR-1.3B 
from which Carballo-bloom-1.3B was derived, and Cerebras-GPT-1.3B, 
from which Carballo-cerebras-1.3B was derived.

Table \ref{tab:taskeval} summarizes the results obtained. The first two rows show the evaluation of the models introduced in this work, while the last five rows provide the results for the baseline models. Our two models clearly outperform the baseline in the OpenBookQA task, although the results are less clear in the remaining cases. The evaluation of Carballo-cerebras-1.3B for the Belebele task demonstrates a competitive achievement, while Carballo-bloom-1.3B shows performance comparable to the best-performing baseline. Notably, however, the results of all models are generally very low, close to a random classification. This seems to show that LLMs with architectures around 1B parameters and without having undergone instruction tuning are not prepared to deal with this type of tasks.

\section{Conclusions}
According to the qualitative evaluation performed, we can conclude that our LLMs are capable of generating high-quality and semantically coherent text in Galician. However, we have observed that automatic evaluation through few-shot learning based on specific instruction tasks may not be suitable for architectural models with approximately 1 billion parameters that have not been instructed for those tasks. 

Furthermore, our findings indicate that Carballo-cerebras-1.3B, based on a monolingual English model (Cerebras-GPT-1.3B), performs better than Carballo-bloom-1.3B, which is based on a multilingual model (Bloom-1b7), in terms of the formal quality of the text generated. As both models were trained on the same intermediate trilingual English/Catalan/Spanish model, this difference in performance could be due to the textual quality of the underlying base model.

Looking ahead, our future work will focus on, at least, two main research lines: firstly, we plan to develop models with larger architectures, albeit not excessively large, and train them on more extensive corpora, encompassing not only Galician but also Portuguese and other neighboring languages to facilitate multilingualism with closed languages. Taking into account these multilingual models with a high percentage of typologically close languages, we will evaluate whether Galician improves in a multilingual context with respect to a monolingual one, and we will investigate whether there are tasks for which monolingual models are better suited than multilingual ones. Secondly, we aim to create models specifically instructed in Galician upon the preparation of instruction datasets in the same language, and we will try to carry out these experiments in coordination with other groups working with similar instruction datasets for the rest of Iberian languages. Hopefully, our work will not only contribute to further advancements in language technologies in the domain of Galician language modeling, but also for other Iberian languages.

\section*{Acknowledgments}
This publication was produced within the framework of the Nós Project, which is funded by the Spanish Ministry of Economic Affairs and Digital Transformation and by the Recovery, Transformation, and Resilience Plan - Funded by the European Union - NextGenerationEU, with reference 2022/TL22/00215336, and by the Xunta de Galicia through the collaboration agreements signed in with the University of Santiago de Compostela in 2021 and 2022. 

Additionally, the authors of this article received funding from MCIN/AEI/10.13039/501100011033 and the European Union Next Generation EU/PRTR (TED2021-130295B-C33), the Galician Government (ERDF 2014-2020: Call ED431G 2019/04, and ED431F 2021/01), by MCIN/AEI/10.13039/501100011033 (grants with references PID2021-128811OA-I00, PLEC2021-007662, and TED2021-130295B-C33, the latter also funded by the European Union Next Generation EU/PRTR), a Ramón y Cajal grant (RYC2019-028473-I), and a Juan de la Cierva Grant (JDC2022-049433-I) funded by MCIN/AEI/10.13039/501100011033 and the European Union Next Generation EU/PRTR. 

 We are grateful to CESGA (Centro de Supercomputación de Galicia) for allowing us access to their infrastructure to carry out the experiments.





\bibliographystyle{fullname}
\bibliography{main.bib}

\appendix

\section{Appendix A: Translation of the example of an evaluation context}
\label{sec:anex}

\begin{itemize}
  \item \textbf{Context:} \textit{Universities will be able to hire teachers of modern languages for their lectureships as part of }
  \item \textbf{Authentic continuation:} \textit{the staff of the responsible departments. The statutes of each university will determine the characteristics of this staff. Their teaching obligations will be fixed in the respective contracts, by virtue of the teaching needs of the department, being at most, those of a collaborating teacher.}
  \item \textbf{Generated continuation:} \textit{their teaching program. The hiring of these professors must be approved by the corresponding university, which may establish the specific conditions and requirements for candidates for lecturer or scholarship based on the needs of the academic institution in which the}
\end{itemize}

\end{document}